\documentclass[11pt,a4paper]{article}
\usepackage[hyperref]{acl2021}
\usepackage{times}
\usepackage{latexsym}

\usepackage{microtype}
\usepackage{url}
\usepackage{times}
\usepackage{latexsym}
\usepackage{tabularx}
\usepackage{graphicx}
\usepackage{booktabs}
\usepackage{amsfonts}       
\usepackage{nicefrac}       
\usepackage{microtype}      
\usepackage{multirow}
\usepackage{xcolor}
\usepackage{caption}
\usepackage{amsmath}
\usepackage{float} 
\usepackage{ragged2e}

\usepackage{enumitem}
\mathchardef\mhyphen="2D

\usepackage{appendix}
\usepackage{etoolbox}
\makeatletter
\patchcmd{\hyper@makecurrent}{%
    \ifx\Hy@param\Hy@chapterstring
        \let\Hy@param\Hy@chapapp
    \fi
}{%
    \iftoggle{inappendix}{
        \@checkappendixparam{chapter}%
        \@checkappendixparam{section}%
        \@checkappendixparam{subsection}%
        \@checkappendixparam{subsubsection}%
        \@checkappendixparam{paragraph}%
        \@checkappendixparam{subparagraph}%
    }{}%
}{}{\errmessage{failed to patch}}

\newcommand*{\@checkappendixparam}[1]{%
    \def\@checkappendixparamtmp{#1}%
    \ifx\Hy@param\@checkappendixparamtmp
        \let\Hy@param\Hy@appendixstring
    \fi
}
\makeatletter

\newtoggle{inappendix}
\togglefalse{inappendix}

\apptocmd{\appendix}{\toggletrue{inappendix}}{}{\errmessage{failed to patch}}
\apptocmd{\subappendices}{\toggletrue{inappendix}}{}{\errmessage{failed to patch}}

\usepackage{contour}
\usepackage[normalem]{ulem}

\contourlength{0.8pt}

\newcommand{\nuline}[1]{%
  \uline{\phantom{#1}}%
  \llap{\contour{white}{#1}}%
}

\DeclareMathOperator*{\argmax}{argmax}

\aclfinalcopy 

\title{Prompting Contrastive Explanations for Commonsense Reasoning Tasks}

\author{Bhargavi Paranjape$^{\dagger*}$ \quad Julian Michael$^{\dagger}$ \quad Marjan Ghazvininejad$^{*}$ \\ \bf Luke Zettlemoyer$^{\dagger*}$ \quad \bf Hannaneh Hajishirzi$^{\dagger\epsilon}$ \\ $^{\dagger}$Allen School of Computer Science \& Engineering, University of Washington, Seattle, WA \\  $^{\epsilon}$Allen Institute of Artificial Intelligence, Seattle \\ $^{*}$Facebook AI \\ 
{\tt \{bparan,julianjm,lsz,hannaneh\}@cs.washington.edu}}

\date{}

\begin{document}
\maketitle
\begin{abstract}
  Many commonsense reasoning NLP tasks involve choosing between one or more possible answers to a question or prompt based on knowledge that is often implicit.
  Large pretrained language models (PLMs) can achieve near-human performance on such tasks, while providing little human-interpretable evidence of the underlying reasoning they use.
  In this work, we show how to use these same models to generate such evidence: inspired by the contrastive nature of human explanations, we use PLMs to complete explanation prompts which contrast alternatives according to the key attribute(s) required to justify the correct answer (for example, \textit{\nuline{peanuts} are usually salty while \nuline{raisins} are sweet}). Conditioning model decisions on these explanations improves performance on two commonsense reasoning benchmarks, as compared to previous non-contrastive alternatives. These explanations are also judged by humans to be more relevant for solving the task, and facilitate a novel method to evaluate explanation faithfulness.
\end{abstract}

\section{Introduction}

Pretrained Language Models (PLMs) \cite{raffel2019exploring, lewis-etal-2020-bart, radford2019language, brown2020language} have been shown to encode substantial amounts of knowledge in their parameters \cite{petroni2019language, talmor2020olmpics, roberts2020much} and have achieved impressive performance on commonsense reasoning (CSR) tasks without the use of external knowledge \cite{trinh2018simple, yang2020g}. 
However, these models provide little human-interpretable evidence of the intermediate commonsense knowledge or reasoning they use, and have been observed to overly rely on superficial dataset artifacts \cite{poliak2018hypothesis, geva2019we}. To overcome this limitation, recent work has shown that PLMs can explain themselves by {\it generating} free-form natural language explanations of their reasoning patterns \cite{rajani2019explain, camburu2018snli, narang2020wt5}. 
However, the space of possible free-form explanations is incredibly large, inherently ambiguous, and difficult to annotate or evaluate \cite{wiegreffe2020measuring, latcinnik2020explaining}. Furthermore, quantifying the model's dependence on free-form explanations is also challenging \cite{camburu-etal-2020-make}.
We address these challenges by proposing an unsupervised method that uses contrastive prompts, which require the model to \emph{explicitly} contrast different possible answers in its explanation (Table~\ref{tab:teaser_figure}).

\begin{table}[]
    \centering
    \small
    \begin{tabular}{l}
    \toprule
        i) I picked up a bag of \textcolor{red}{peanuts} and \textcolor{blue}{raisins} for a snack. \\ I wanted a sweeter snack out so I ate the \_\_ for now. \\ 
        {\it Contrastive Expl. -  Peanuts are salty while raisins tend}\\{\it  to be sweet.}\\
         \midrule
        ii) The geese prefer to nest in the \textcolor{blue}{fields} rather than the \\ 
\textcolor{red}{forests} because in the $\_\_$ predators 
        are more hidden. \\
        \textit{Contrastive Expl. - Forests are denser than fields}\\
    \bottomrule
    \end{tabular}
    \caption{Examples of Winograd Schema Instances where the correct and incorrect answer choices are highlighted in blue and red respectively. Choices are \emph{contrasted} along attributes like taste (for i) and density of vegetation (for ii) by humans to explain why they prefer some answer choice.} 
    \label{tab:teaser_figure}
\end{table}

Our approach is based on a key observation: 
Many commonsense reasoning tasks require the comparison or contrast of plausible alternatives along a distinguishing attribute. For instance, in Table~\ref{tab:teaser_figure}, the differentiating attributes for the two answer choices maybe taste (for i) and vegetation density (for ii). 
People commonly use contrastive explanations to explain their reasoning~\citep{miller2018contrastive}. 
Rather than asking ``Why P?", they ask ``Why P rather than Q?", where Q may be implicit from the context. For example, instead of justifying why raisins are the appropriate choice, people tend to explain why they are more likely than peanuts.
\citet{miller2018contrastive} also argues that such contrastive explanations are computationally efficient, as they
only require focusing on the limited set of reasons that might make one answer more likely than the other instead of exhaustively enumerating all possible reasons for an answer.
For instance, the raisin's taste (not its size, temperature, etc.) in Table~\ref{tab:teaser_figure} is adequate to explain why it is the best answer.

 \begin{table*}[ht!]
    \centering
\begin{small}
    \begin{tabular}{l|l}
        \toprule
        \bf Dataset Instance & \bf Human-Authored Contrastive Explanation\\
        \midrule
        \bf{Winograd Schema} & \\
        1. The \textcolor{red}{party} was more  interesting and uplifing than the   & $\circ$ Parties are for celebrating while funerals are for mourning \\
        \textcolor{blue}{funeral} because the $\_\_$ was rigid. & $\circ$ People wear colorful clothes at parties and black at funerals \\
        2. The geese prefer to nest in the \textcolor{red}{fields} rather than the   & $\circ$ Forests are dense while fields are sparse \\
        \textcolor{blue}{forests} because in the $\_$ predators are more hidden.  &  $\circ$ Forests have more predators than fields.\\
        \midrule
        \bf{PIQA}  & \\
        1. How do you get strong hamstrings? \_\_ & $\circ$ Hamstrings are located in the legs while biceps are  located in \\
        (a) \textcolor{red}{work out your upper body} (b) \textcolor{blue}{work out your legs} & the upper body \\
        2. How do you flood a room? \_\_	& $\circ$ Filling it with objects can clutter a room while filling it  \\
        (a) \textcolor{red}{fill it with objects} (b) \textcolor{blue}{fill it with water} &  with water floods the room. \\
        \bottomrule
    \end{tabular}
    \end{small}
    \caption{Examples of commonsense tasks that can be explained using contrastive language and some contrastive explanations authored by in-house annotators. The \textcolor{blue}{Fact} and \textcolor{red}{Foil} are marked in the input.}
    \label{tab:examples_contrastive}
\end{table*}

Our goal is to enable PLMs that explain their predictions to similarly benefit from such constraints. 
 We develop a small set of contrastive generation prompts that can be in-filled by a PLM such as T5 \cite{raffel2019exploring} or BART \cite{lewis-etal-2020-bart} (see Table~\ref{tab:contrastive_patterns}). These templates are designed to cover a multitude of language patterns used by humans to compare and contrast entities. 
 Another PLM then conditions on both the original input and the generated contrastive explanation, to predict the final answer.
 This approach is inspired by \citet{shwartz2020unsupervised}, who also use textual prompts to query the PLM with clarification questions. However, their prompts are generic while we prompt for  instance-specific information. 

Our approach shows quantitative improvements in task performance over two existing methods for model explainability \citep{shwartz2020unsupervised, latcinnik2020explaining}, for
two commonsense reasoning tasks: the Winograd Schema Challenge \cite{levesque2012winograd} and multiple-choice question answering about physical commonsense
\cite{Bisk2020}.
Our gains in the zero-shot setting are especially notable, outperforming the best reported results on publicly available PLMs and improving over \citet{shwartz2020unsupervised} by up to 11\%. 
We also show, through human evaluations, that contrastive explanations are deemed more useful for solving the original task compared to generic clarification questions. 
Finally, contrastive explanations can be semantically perturbed to quantify the model's dependence on them by flipping the contrast in the explanation to support the foil, facilitating quantification of model faithfulness.\footnote{Code is available at \url{https://github.com/bhargaviparanjape/RAG-X}}


\section{Related Work}
Models that rationalize their decisions by extracting a contiguous subsequence of the input as an explanation \cite{lei2016rationalizing, deyoung-etal-2020-eraser, paranjape-etal-2020-information} are inadequate in explaining commonsense reasoning tasks that require knowledge that is implicit in the input. Such tasks necessitate PLMs to rely on embedded parametric knowledge. 
Recent work use free-form textual explanations to generate explanations for commonsense reasoning tasks like SNLI \cite{camburu2018snli}, Winograd Schemas \cite{zhang2020winowhy} and CommonsenseQA \cite{rajani-etal-2019-explain} through explicit human supervision, which are inherently ambiguous, incomplete and consequently, expensive to collect and evaluate on \cite{camburu2019make, camburu2019can, deyoung-etal-2020-eraser}. Most recently, ~ \citet{latcinnik2020explaining} use an unsupervised approach to generate free-form explanations as sequences of tokens that are not well-formed sentences. 
In contrast, our method uses specialized prompts to generate well-formed human-interpretable explanations without any additional supervision. 

 
Specialized prompts have been shown useful for extracting knowledge from PLMs in a targeted manner
\cite{petroni2020context, richardson2020does, talmor2020olmpics, donahue2020enabling, lin2019commongen} and improving performance on downstream tasks \cite{brown2020language, shin2020autoprompt}. 
Most relevant to our work is the self-talk model 
of \citet{shwartz2020unsupervised}, an unsupervised approach using a fixed set of clarification questions as prompts to elicit knowledge from PLMs for commonsense reasoning tasks.
Our work differs by focusing specifically on contrastive PLM prompts, which we find further improve performance by eliciting explanations which are highly relevant to the classification decision (Section~\ref{sec:experimental-results}).

Our approach to contrastive reasoning is also closely related to \textit{counterfactuals}, which can be used to give contrastive explanations, i.e., answers to ``Why P rather than Q?'', by providing a counterfactual case in which Q would have held. \citet{ross2020explaining} use this idea to generate contrastive explanations, while it has also been used for evaluation \citep{gardner2020evaluating} and training \citep{kaushik2019learning} with the aim of addressing model robustness.
Most of this work explicitly constructs counterfactual cases by perturbing the input data of a task in order to produce changes in the output label.
In contrast, we do not construct counterfactual \textit{inputs}, but aim to explicitly represent counterfactual \textit{knowledge}: a contrast between the fact P and foil Q that, were it hypothetically \textit{reversed}, would change the output label. We include an evaluation of our models on this question in Section~\ref{sec:faithfulness_improvements}.

\section{Contrastive Explanations}
We present the theory of contrastive explanations adopted in this work (Section~\ref{sec:def}) and the intuition behind using them for commonsense reasoning tasks (Section~\ref{sec:cs}).
\subsection{Definition and Motivation\label{sec:def}}
A contrastive explanation is generally defined as an answer to a counterfactual question of the form ``Why P rather than Q?" for two potential hypotheses $P$ and $Q$ that can follow from some event $E$. It explains why some \emph{fact} $P$ occurred instead of some \emph{foil} $Q$, where $Q$ can be implicit \cite{hesslow1988problem, lipton1990contrastive, miller2019explanation}.  
A good contrastive explanation points to differences between the fact and foil with regard to certain attributes, not just conveying that the fact has a certain attribute. 
Table \ref{tab:teaser_figure} shows examples of contrastive explanations that differentiate between peanuts and raisins (on the basis of taste) or forests and fields (on the basis of vegetation densities) to explain the more probable answers to Winograd Schema instances.

Previous studies~\cite{miller2019explanation} in  philosophy, psychology, and cognitive science show that humans use such contrastive explanations when explaining their decisions to each other. Importantly, \citet{miller2018contrastive} also argues that contrastive explanations are computationally efficient -- exhaustively describing all causes for the occurrence of an event $P$ is harder than only enlisting causes for why another event $Q$ did not occur instead of $P$. 


\subsection{Contrastive Explanations for Commonsense Reasoning Tasks \label{sec:cs}}
Many recently proposed commonsense reasoning tasks are framed in a multiple-choice format that facilitates contrastive explanation (see \autoref{tab:examples_contrastive}). In this study, we focus on the following two tasks.

The \textbf{Winograd Schema Challenge} \citep[WSC]{levesque2012winograd} is a pronoun coreference resolution task designed as a hard benchmark for evaluating everyday knowledge and commonsense reasoning \cite{zhang2020winowhy}. For instance, in the sentence ``The city councilmen refused the demonstrators a permit because they feared violence," the pronoun \textit{they} must be disambiguated between fact (\textit{the city councilmen}) and foil (\textit{the demonstrators}). Both fact and foil are explicit in such sentences.

The \textbf{Physical Interaction Question Answering} \cite[PIQA]{Bisk2020} challenge is designed to test knowledge of physical commonsense. PIQA requires choosing between which one of two \textit{solutions} is a better way of achieving a \textit{goal} posed as a question (see \autoref{tab:examples_contrastive}). PIQA questions relate to physical properties of entities, their affordances, and how they can be manipulated.
The fact and foil are explicit in the two solutions, which typically differ from one another by a short noun phrase. 

To validate our intuition that contrastive reasoning is instrumental in these tasks, we performed a pilot study with 10 annotators over 100 commonsense questions from Winogrande and PIQA.
We instructed them to answer the questions and explain their reasoning, but gave no specific instructions about what the explanations should look like. 
Examples are shown in \autoref{tab:examples_contrastive}.
In 76\% of Winogrande and 64\% of PIQA examples, annotators explicitly contrasted the fact and foil.
The frequent use of certain phrase structures, like \textit{P are \_\_ while Q are \_\_}, strongly informed our method for generating them (\autoref{sec:modeling}).


\begin{table*}[ht!]
    \centering
    \begin{small}
    \begin{tabular}{ll}
    \toprule
    \bf Prompt Pattern & \bf Commonsense Example \& Model Generated Explanation \\
    \midrule
    \bf Personal Characteristics \\
    $\implies P$ likes/likes to $\_$ while $Q$ likes/likes to $\_$ & \textcolor{red}{Megan} said it would be liberating to go out without makeup like \\
    $P$ likes/likes to $\_$ while $Q$  does not like/like to $\_$ &  \textcolor{blue}{Elena} does since \_\_ never wore makeup \\
    $P$ prefers/prefers to $\_$ while $Q$ prefers $\_$ &  {\tt Explanation: Elena likes to \underline{be natural} while }\\
    $Q$ prefers $\_$ while $P$ does not prefer/prefer to $\_$ & {\tt Megan likes to \underline{wear lipstick} } \\
    $Q$ thinks $\_$ while $P$ thinks/does not think  $\_$ & \\
    \midrule
    \bf Object Characteristics \\
    $P$ is taller/shorter/smaller/larger/slower/faster than $Q$  & How to tie pieces of paper together? \_\_ \\
    $\implies P$ is/are $\_$ while/but/however $Q$ is/are $\_$ & (a) Thread \textcolor{red}{ruler} through the holes \\
    $Q$ has/have $\_$ while/but/however $P$ has/have $\_$   & (b) Thread \textcolor{blue}{ribbon} through the holes \\
    P has/have more/less $\_$ than Q& {\tt Explanation: Ruler is \underline{hard} while a ribbon is }\\
    P is/are $\_$ than Q & {\tt \underline{flexible}}\\
    \midrule
    \bf Spatial/Temporal Contrast \\
    $\implies P$ is inside/outside/above/below $Q$ & \textcolor{red}{Emily} looked up and saw \textcolor{blue}{Patricia} racing by overhead. \_\_ was on the \\
    $\_$ is closer to $P$ and farther away from $Q$  &  ramp. \\
    $P$ is to the right/left of $Q$  & {\tt Explanation: Emily is below Patricia}\\
    $Q$ takes longer to $\_$ than $P$   &  \\
    \midrule
    \bf Use cases and causes \\
    $P$ is used for $\_$ $Q$ & To prepare the puff pastry for your pie, line a baking sheet with \\
    $P$ is used to do $Q$ $\_$  & parchment. Then \_\_ 	\\
    $\implies P$ is used for/to/in $\_$ while $Q$ is used for/to/in $\_$  & (a) Unroll the pastry, lay it over \textcolor{blue}{baking twine}. \\
    $Q$ is used $\_$ while $P$ is used $\_$ & (b) Unroll the pastry, lay it over \textcolor{red}{fishing line}. \\
    $Q$ because $\_$ while $P$ because $\_$  & {\tt Explanation: Baking twine is used in }\\
    $Q$ can cause $\_$ while $P$ results in $\_$  & {\tt \underline{baking} while fishing line is used in \underline{fishing}}\\
    \bottomrule
    \end{tabular}
    \end{small}
    \caption{Contrastive Patterns and Examples of outputs generated by the T5-large model. The pattern the PLM completes are marked $\implies$.}
    \label{tab:contrastive_patterns}
\end{table*}

\section{Our Approach}
\label{sec:modeling}
\begin{figure*}[t]
    \centering
    \includegraphics[scale=0.36]{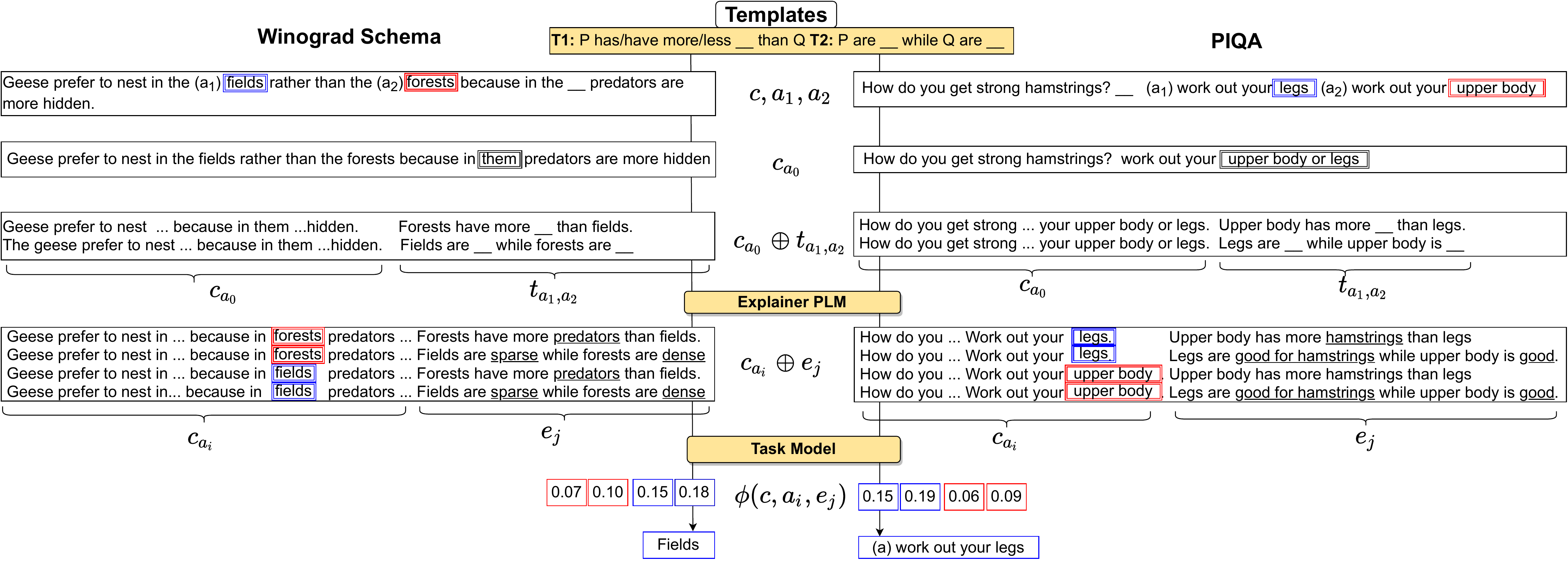}
    \caption{(1) A commonsense reasoning instance $(c,a_1,a_2)$ is converted into a custom prompt $c_{a_0} \oplus t_{a_1,a_2}$ as input for the explainer PLM (2) The combination of input and explanation ($c_{a_i} \oplus e_j$) is used by task model to score $a_i \forall i \forall j$. For $a_1$ and $a_2$, scores are aggregated over templates. }
    \label{fig:model_design}
\end{figure*}

We assume the input to a commonsense reasoning problem consists of a textual context $c$
which contains a placeholder \_, 
and two marked answer choices $a_1$ and $a_2$ corresponding to the fact and foil (Table \ref{tab:examples_contrastive}, left column).
Let $c_x$ denote substitution of $x$ for the placeholder in $c$. The task is to predict 
whether $c_{a_1}$ or $c_{a_2}$ is more likely to be true, i.e., whether $a_1$ or $a_2$ best completes the context.

Our approach has two stages:
First, an {\bf Explainer PLM} $P_{expln}$ generates contrastive explanations (Section~\ref{sec:generation}) by infilling preset {\it contrastive templates} (Sec.~\ref{sec:templates}) on the basis of $c$, $a_1$, and $a_2$.
Then, a {\bf Task Model} $P_{LM}$ selects the correct answer conditioned on both the context and the generated explanations (Sec.~\ref{sec:task_model}).



\subsection{Contrastive Templates} \label{sec:templates}
We develop a list of contrastive templates on the basis of an annotation study.
For ~250 instances from Winogrande and PIQA, we asked three annotators to explain why one answer is more likely than the other.
We manually examined these explanations and abstracted them into templates containing at least two placeholders: two for the fact and foil being contrasted, and possibly more corresponding to the properties they are being contrasted on. For instance, {\it peanuts are salty while raisins are sweet} becomes {\it Q are \_ while P are \_}. We retained templates used by annotators at least 10 times. \autoref{tab:contrastive_patterns} shows several examples.
A template is converted into an explanation by replacing placeholders for the fact and foil with answers $a_1$ and $a_2$ and the remaining placeholders with the appropriate contrastive information.

We evaluate the quality and coverage of our templates with another round of human evaluation. For 100 WSC and PIQA examples, we ask three annotators to either write contrastive explanations using one or more of the templates, or indicate that none of the them were appropriate. Annotators used the templates in over 82\% of cases, indicating high coverage for the tasks we study. 

\subsection{Generating Explanations \label{sec:generation}}
Let $t$ denote a contrastive template.
We write $t_{a_1,a_2}$ to denote the customization of $t$ to an input by filling its marked placeholders for fact and foil with the answer choices. For instance, in Figure~\ref{fig:model_design}, the template {\it P are \_ while Q are \_} is customized to {\it Fields are \_ while forests are \_}.\footnote{In practice, we randomize the order of $a_1$ and $a_2$ when customizing the template.}
A full explanation may be produced by filling the remaining gaps in $t_{a_1,a_2}$ by leveraging an infilling language model, the explainer $P_{expln}$.

\newcommand*{\expl}{\ensuremath{\mathsf{e}}}

We first construct a neutral context $c_{a_0}$ by filling $c$'s placeholder with a task-specific neutral answer that does not indicate if $a_1$ or $a_2$ is correct. For Winogrande Schemas, $c_{a_0}$ is constructed using the ambiguous pronoun in $c$ ({\it them} in Figure~\ref{fig:model_design}). For PIQA, $c_{a_0}$ is constructed as $``c \oplus a_1\text{ or }a_2$'', where $\oplus$ is string concatenation, e.g., {\it upper body or legs} in Figure~\ref{fig:model_design} (More dataset-specific details are in Section~\ref{exp:task_details}).
We then prepend $c_{a_0}$ to the customized template $t_{a_1,a_2}$ and use it as input to the infilling language model to fill in the remaining gaps in the template. We use the maximum likelihood candidate phrases from top-K decoding to transform the template into a full explanation $\expl$.

We use a list of templates $t_1, \dots, t_n$ to generate a list of candidate explanations $\expl_1, \dots, \expl_n$ for each input, which are all fed into the task model. We also use some task-specific heuristics to reduce the number of prompts for each example, detailed in Appendix~\ref{apdx:contrastive_templates_all}.

\subsection{Task Model \label{sec:task_model}}

\newcommand*{\prob}{\mathsf{P}}
Given the context and answer choices ($c, a_1, a_2$) and a list of explanations $\expl_1, \dots, \expl_n$, the second stage of our pipeline is a binary classifier between $a_1$ and $a_2$ which marginalizes over the explanations.
We first assign a score to each answer $a \in \{a_1, a_2\}$ and explanation $\expl \in \{\expl_1, ..., \expl_n\}$:
\[ \phi(c,a,\expl) = \frac{1}{k} \log \prob_\text{LM}(c_a \oplus \expl), \] 
where $c_a$ denotes the substitution of $a$ into $c$, $\prob_\text{LM}$ is string probability under the task  language model, and $k$ is the string length of $c_a \oplus \expl$.
We use $\phi$ as input to a logistic regression classifier which marginalizes over explanations:
\[\prob(a \mid c, a_1, a_2) = \frac{\sum_i^n e^{\phi(c, a, \expl_i)}}{Z}, \]
where $Z$ is a normalizer over $a_1$ and $a_2$.
At initialization, $\phi$ uses a pretrained language model, and we fine-tune it to minimize the cross-entropy loss of \(\prob(a^* \mid c, a_1, a_2)\), where $a^*$ is the correct answer.
We do not fine-tune the explainer PLM since the top-K beam decoding is a discrete operation that is hard to backpropagate through.
In the zero-shot setting (where the task PLM is not fine-tuned) and during inference, the answer is predicted by aggregating scores assigned to an answer by all $n$ explanations: $\argmax_{a_i} \sum_j \phi( c, a_i,\expl_j)$.

\section{Experimental Setup}
\subsection{Baselines}\label{sec:baselines}

\paragraph{Context-Only}
We experiment with a baseline that does not condition on explanations at all. Here, 
\[ \phi(a, c) = \frac{1}{k} \log \prob_\text{LM}(c_a), \]
and gold answer is $\argmax_{a_i} \phi(a_i, c)$

\paragraph{Unconstrained Generation}
\citet{latcinnik2020explaining} generate explanations from a PLM by beam-decoding a free-form sequence termed a {\it hypothesis} which is then used by a classifier to solve the task. The model is trained end-to-end and loss terms are added to encourage the hypothesis to sound natural. Explanation generation is otherwise unconstrained. For fair comparison with our approach, we do not fine-tune the explainer PLM (more details are in Appendix~\ref{apdx:hyperparameters}).

\paragraph{Self-Talk}
\citet{shwartz2020unsupervised} propose an unsupervised model that uses a PLM as the answer scorer and a (possibly different) PLM as a knowledge source, similar to our framework. They formulate the process of obtaining relevant knowledge as {\it self-talk} with the
following steps: 1) completing clarification question prefixes such as ``what is the definition of ..." conditioned on input context, 2) generating their corresponding answers (clarifications), and 3) conditioning on the clarification questions and answers to make predictions. The key difference between their approach and ours is in the choice of prompts for the PLM, and the kinds of knowledge the prompts seek. While \citet{shwartz2020unsupervised} draw inspiration from inquiry-based discovery learning \cite{bruner1961act}, we target contrastive reasoning.  

\subsection{Implementation details} \label{exp:task_details}
We use BART-Large \cite{lewis-etal-2020-bart} and T5 \cite{raffel2019exploring} as the explainer PLMs. Hyperparameters for infilling are given in Appendix~\ref{apdx:hyperparameters}.  
For a fair comparison of all models, we use  GPT2-XL \cite{radford2019language} as the task model that estimates $\phi(c, a, \expl)$. GPT2-XL is the best performing PLM used by \citet{shwartz2020unsupervised} for WSC and PIQA tasks. Hyperparameter details about finetuning are given in Appendix~\ref{apdx:hyperparameters}. 
We describe dataset specific modifications made to create $c_{a_0}, c_{a_1},$ and $c_{a_2}$ in Section~\ref{sec:generation}. 

\begin{table*}[h]
    \centering
    \begin{small}
    \begin{tabular}{lcc|cccccc}
\toprule
& \bf Explainer & 	\bf Task model& 	\bf WGRD && 	\bf PIQA&& 	\bf WSC& 	\bf WGND \\
& \bf PLM (\# Params) & & \bf ZS & \bf FT & \bf ZS & \bf FT & \bf ZS  & \bf ZS 			\\
\midrule		
1. Context-only & GPT2-XL (1.5B) & GPT2-XL& 	54.8& 77.9  & 62.6 & 80.1& 	61.5& 	60.0 \\
2. Unconstrained & GPT2-XL& & 	54.9& 77.8& 	63.9& 80.7 & 	61.4& 	60.0\\
3. Self-Talk & GPT2-XL	& & 	55.1& 78.4 & 	69.5& 82.3 & 	62.0& 	61.3\\
\midrule
4. Contrastive & BART-Large(680M) & & 		56.8& 	78.9 & 	71.8& 82.8& 	63.2& 	62.9\\
5. \hspace{8pt}(Ours) &  T5-Large (770M) & & 		59.2& 79.1 & 	72.5& 83.5& 	63.5& 	63.2\\
6.&  T5-11B(11B)	& & 	\bf 60.3& 	\bf 79.6& \bf 73.4 &	\bf 83.9&	\bf 64.1& 	\bf 63.5\\
\bottomrule
    \end{tabular}
    \end{small}
    \caption{Test set accuracy on Winogrande (WGRD), PIQA, WSC and Winogender (WGND). ZS is Zero-shot models while FT is fine-tuned models. WSC and Winogender don't have training data for finetuning. Across all our models, the task model is GPT2-XL for fair comparison with \cite{shwartz2020unsupervised} and to make finetuning tractable.}
    \label{tab:final_results}
\end{table*}

\paragraph{Winograd Schema Challenge (WSC) } 
We experiment on  (i) the SuperGLUE \cite{wang2019superglue} version of the WSC  consisting of 285 examples of anaphora (pronoun) resolution; (ii) Winogrande (WGRD) \cite{sakaguchi2020winogrande}, a large scale crowdsourced version of the WSC; and (iii) WINOGENDER (WGND), a diagnostic dataset created to measure gender bias in models for ambiguous pronoun resolution \cite{rudinger2018gender}.

Each instance provides two answer choices, which we use directly as $a_1$ and $a_2$. 
For the neutral answer $c_{a_0}$, we use the sentence with the original ambiguous pronoun. Since Winogrande has a blank space \_ for the answer, we replace it with the most likely pronoun under a masked language model (BERT), following \citet{shwartz2020unsupervised}. $c_{a_1}, c_{a_2}$ are obtained by replacing the blank space or pronoun with the answer choice.

\vspace{-0.1cm}
\paragraph{Physical Interaction Question Answering (PIQA) \cite{Bisk2020}}
PIQA provides two answer choices which mostly vary from each other on a substring (e.g., ``work out your [upper body]/[legs]''). We use these differing substrings as $a_1$=legs and $a_2$=upper body.
For the neutral answer $a_0$, we combine the answers into ``$a_1$ or $a_2$'' (upper body or legs). 
In the cases where $a_1$ or $a_2$ is longer than 2 words, we include an \textit{or} between the full answers. More details and examples are presented in Appendix~\ref{apdx:contrastive_templates_all}. We use question-answer pairs for $c_{a_1}$ and $c_{a_2}$.





\section{Experimental Results}
\label{sec:experimental-results}
In this section, we present an extensive evaluation of our approach, demonstrating performance gains
which are independently verified by human judges.

\begin{table*}
	\begin{minipage}{0.65\linewidth}
		\centering
    \begin{small}
    \begin{tabular}{@{}l|cc|cc|cc@{}}
    \toprule
        \bf Metric & \multicolumn{2}{c}{\bf Self-Talk (Reported) } & \multicolumn{2}{c}{\bf Self-Talk} & \multicolumn{2}{c}{\bf Contrastive} \\
        & \bf WGRD & \bf  PIQA & \bf  WGRD & \bf  PIQA & \bf WGRD & \bf PIQA \\
        \midrule 
Relevant &	68 &	60 &	70.4 &	61.7&	73.1&	70.7 \\
Factual &	46&	42	&40.8	&38.8&	43.0&	39.4 \\
Helpful &	24&	26&	22.5&	27.7&	42.8&	32.8 \\
Grammatical	& 87.2&	87.2&	87.5&	87.5&	83.5&	83.5 \\
Flips       & NA&	NA&	NA&	NA&	66.9 &	59.4 \\
    \bottomrule
    \end{tabular}
    \end{small}
\caption{Human Evaluation Results on Winogrande(WGRD) and PIQA. Reported human evaluation results on Self-talk are different from ours, which can be because of moderate levels of agreement (Fleiss Kappa $\kappa$ = 0.43).
Grammatiality is judged together for all datasets following \cite{shwartz2020unsupervised}. Only contrastive explanations can be flipped.} \label{tab:human_eval}
	\end{minipage}\hfill
	\begin{minipage}{0.30\linewidth}
		\centering
		\includegraphics[scale=0.32]{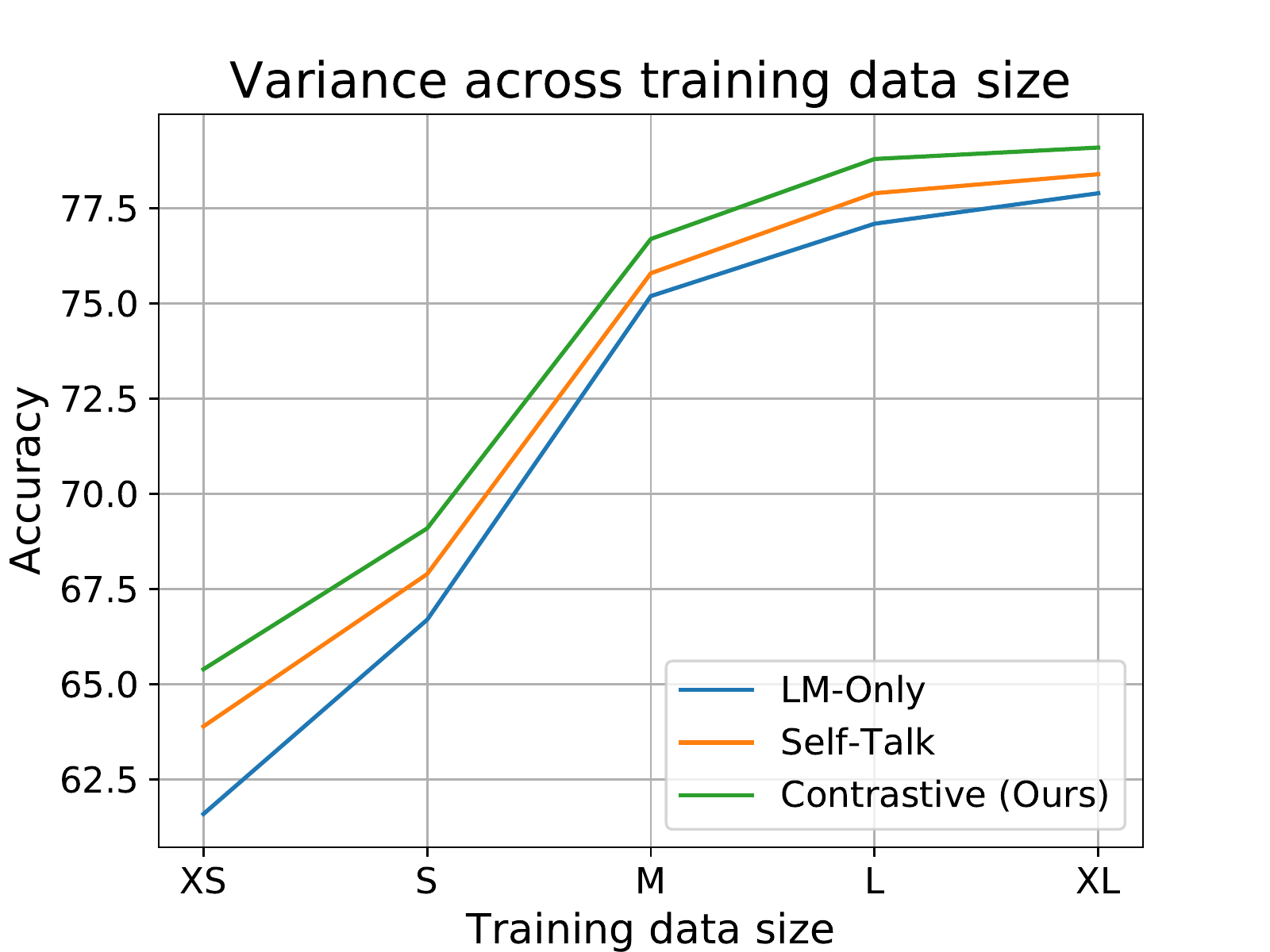}
		\captionof{figure}{Performance variation in the finetuning setting across different sizes of Winogrande training data.}
    \label{fig:training_size_influence}
	\end{minipage}
\end{table*}

\begin{table*}[]
    \centering
    \small
    \begin{tabular}{l|l|l|l}
    \toprule
    \bf Example & \bf Unconstrained & \bf Self-Talk & \bf Contrastive \\
    \midrule
 (i) \textcolor{red}{Ian} volunteered to eat \textcolor{blue}{Dennis's} & Dennis's menudo  & What are the properties & Dennis is a \underline{vegetarian} while  \\  
 menudo after already having a  & was disgusting. & of \underline{a menudo? A menudo} &  Ian is a \underline{carnivore.} Dennis has  \\ 
  bowl because he despised  &  	&   \underline{is made from the}  &  \underline{menudo} while Ian has \underline{volunteered} \\
 eating intestine. &&  \underline{intestines of a pig} &  \underline{to eat Denni's menudo.}\\
\midrule          
(i) The \textcolor{red}{GPS} and \textcolor{blue}{map} helped me & because the GPS  & What is \underline{going on here?} & The GPS can \underline{tell me where I am} \\  
navigate home.  I got lost when & and map helped & \underline{The iphone app is not} & but the map can't. \\ 
the it got turned upside down. &  me navigate  &  \underline{working}.	&  The GPS is \underline{right-side-up} while \\
& home. & & the map is \underline{upside down} \\
\midrule 
(ii) I helped my sister find her  & She couldn't wear &What are the properties of & Gold necklace is used \underline{for formal} \\
\textcolor{red}{gold necklace}. She couldn't wear & her woven  & \underline{gold?} The properties of & \underline{occasion} while woven necklace  \\
 her \textcolor{blue}{woven necklace} to the & necklace. &  \underline{gold are listed below.} & is used \underline{for casual occasion.}\\
ball because it was so casual.	&& &	 \\
    \bottomrule
    \end{tabular}
    \caption{Qualitative Examples on Winogrande where contrastive explanations (using T5-11B explainer) improve task performance over baselines.}
    \label{tab:qualitative}
\end{table*}

\begin{table*}
\centering
    \small
    \begin{tabular}{l|cccccc}
\toprule
\bf  Explainer & 	\multicolumn{2}{c}{\bf WGRD} &   \multicolumn{2}{c}{\bf PIQA} & \bf 	WSC& \bf 	WGND\\
\bf PLM & \bf ZS & \bf FT & \bf ZS & \bf FT & \bf ZS  & \bf ZS 			\\
\midrule		
BART-Large & 	53.9 (5.4) & 75.9 (4.0 )& 	66.5 (7.9) & 79.1 (4.6) & 	59.1 (6.9) & 	58.7 (7.1) \\
T5-Large & 56.2 (5.3) & 75.3 (5.0) & 	68.1 (6.5) & 80.2 (4.2) & 60.2 (5.5) & 	59.0 (7.1) \\
T5-11B	& 57.6 (4.5) & 76.1 (4.7) & 	69.5 (5.4) & 81.0 (3.6) & 	61.1 (3.3) & 	59.0 (5.8) \\
\bottomrule
    \end{tabular}
    \caption{Flipped evaluation results for contrastive explanation models. Reporting test accuracy across all datasets. Percent drop in performance as a result of flipping is indicated in parentheses.}
    \label{tab:reversed_results}
\end{table*}

\begin{table}
\centering
\small
\begin{tabular}{lcc}
\toprule
\bf Input    & \bf WGRD & \bf PIQA\\
\midrule		
Fully abstracted   & 63.2 & 54.6\\
Abst. after expl.  & 70.4 & 64.9\\
No abstraction     & 79.1 & 83.5\\


\bottomrule
\end{tabular} 
\caption{Evaluation of fine-tuned T5-Large contrastive models on Winogrande with abstracted answers.}\label{tab:abstractive}
\end{table}


\subsection{Task Performance} \label{sec:task_improvements}
We report task accuracy as a proxy for explanation quality.
Table \ref{tab:final_results} compares the task performance of our model with the baselines defined in Section~\ref{sec:baselines}.  
We observe that generating and conditioning on additional information from PLMs improves performance over just using the original input (Row 1 vs. 2-6).
Using templates to prompt the PLM for specific knowledge is better than unconstrained generation of text (Row 2 vs. 3-6).
Contrastive explanations outperform previous work that use clarification questions in self-talk~\cite{shwartz2020unsupervised}.
The T5-Large explainer already surpasses the results of self-talk despite being smaller than GPT2-XL, demonstrating the impact of using contrastive explanations over clarification questions. 

We also observe that larger explainer PLMs (going from T5-Large to T5-11B) yield higher performance. Our zero-shot results with T5-11B are the highest reported on Winogrande, PIQA and WSC for an open-sourced model.\footnote{The zero-shot SOTA model \cite{brown2020language} uses the 175B parameter GPT-3 model, which would likely also be a stronger explainer for our approach, but we did not have access to it.} 


Finally, our approach gets smaller improvements when finetuning the task model. This suggests that some of the reasoning is still learned implicitly by the task model.
Figure~\ref{fig:training_size_influence} shows task performance with various training data sizes of Winogrande, indicating a larger gap between the Context-Only baseline and our approach when training data is scarce.

\subsection{Human Evaluation}\label{sec:human_evaluation}

\paragraph{Setup}

Following the human evaluation setup of \citet{shwartz2020unsupervised}, we sample up to 50 highest-scoring explanations from PIQA and Winogrande examples which the T5-Large model got correct but the Context-Only baseline failed at. For comparison, we also include explanations from the self-talk model for evaluation.

Crowd workers are presented with a commonsense instance, the correct answer, and an explanation, 
and are asked to judge for: 1) {\it Grammaticality},  whether the explanation is grammatical;
2) {\it Relevance}, whether it's relevant to the topic of the text; 
3) {\it Factual Correctness}, whether it's factually correct or likely true; and 
4) {\it Helpfulness}, whether it adds helpful evidence for the correct answer.
These metrics and definitions follow from \citet{shwartz2020unsupervised} with more details in Appendix~\ref{apdx:human_evaluation}. 
The annotators are also shown the same explanation with fact and foil flipped (details in Section~\ref{sec:faithfulness_improvements}) and are asked to judge if the other answer is more likely than before if they assume the flipped explanation to be hypothetically true.
\vspace{-0.1cm}
\paragraph{Results} Table~\ref{tab:human_eval} shows the results of human evaluation of contrastive and self-talk explanations. 
The contrastive explanations are overwhelmingly preferred over self-talk explanations for relevance, factual correctness and helpfulness. They may be considered less grammatical because of in-filling noise (such as incomplete phrases).
Table~\ref{tab:qualitative} presents some qualitative examples of instances where contrastive explanations improve over all baselines.

\subsection{Analysis}\label{sec:faithfulness_improvements}
We also analyze how much the task model relies on contrastive explanations for its decisions.
\paragraph{Flipping Explanations}
Our choice of contrastive language templates facilitates a novel way to evaluate explanation usefulness in prediction. The contrast in the explanation can be reversed by flipping the position of the fact and the foil in the explanation. 
If the choice between fact and foil actually depends on the contrastive explanation, then the flipped explanation should provide a hypothetical situation where the foil is more likely than the fact. For instance, ``peanuts are salty while raisins are sweet," when switched to ``raisins are sweet while peanuts are salty," may provide evidence that {\it peanuts} is a more likely label for the example in Table~\ref{tab:teaser_figure} (i). 
This may cause a model that uses the explanation to flip its prediction and lead to a drop in accuracy. The magnitude of drop can quantify the extent to which the model relies on the contrast provided in the explanation. In fact, humans also deem the flipped explanation to
imply the opposite label in a majority of cases (Table   \ref{tab:human_eval}), indicating that our contrastive explanations frequently capture contrastive properties that the labels truly rely on.

Table~\ref{tab:reversed_results} shows the flipped evaluation results. We observe declines in accuracy of up to 8\%, indicating that the model does use some contrastive knowledge to reason about the task.
Finetuned models show a smaller decline in accuracy compared to the zero-shot setting. In this case, the task model may be directly fitting the data in lieu of relying on the knowledge conveyed by the explanation.

%
\paragraph{Abstracting Fact and Foil}
Given input context $c$ (consisting of the fact and foil $a_1$, $a_2$) and an explanation $e$, the explainer PLM $P_{expl}$ infills its explanation $e$ on $c$ while the task model $P_{LM}$ conditions on both $c$ and $e$.
We can test the quality of the generated explanations and the task model's reliance on them by forcing the task model to rely on $e$ when information in input $c$ is restricted. One potential way to do so is to scrub the identities of the fact and foil, $a_1$ and $a_2$, from $c$.  

We replace the fact and foil with placeholder tokens to create an abstract context $c'$. For instance, the example in Table~\ref{tab:qualitative} (ii) becomes ``The $\texttt{<mask1>}$ and $\texttt{<mask2>}$ helped me navigate ... down.", where the model must now choose between $\texttt{<mask1>}$ and $\texttt{<mask2>}$.\footnote{More examples of abstracted contexts and explanations are given in the Appendix (Table~\ref{tab:abstractive_example}).}
Running the task model on $c'$ lower-bounds the performance possible without knowing answer identities.
We can now test the relevant answer-based knowledge contrastive contained in the \textit{explanations} by allowing the explanation model to see the original answers in $c$, but then abstracting them out when passing the input context and explanations to the task model. More formally, the task model conditions its decision on $c'$ and $e'$. For Table~\ref{tab:qualitative} (ii) $c'$ and $e'$ are ``The $\texttt{<mask1>}$ and $\texttt{<mask2>}$ helped me navigate ... down." and ``The $\texttt{<mask1>}$ is right-side-up while the $\texttt{<mask2>}$ is upside down."
Since only the explainer PLM is shown answer identities, the task model's decision is conditionally independent of the answer identities given the explanation.


Experiments on Winogrande and PIQA in the fine-tuned setting  (Table~\ref{tab:abstractive}) show that performance improves significantly when the task model conditions on both $c'$ and $e'$ compared to a fully abstracted contrastive baseline that only conditions on $c'$ (from 63.2 to 70.4 for Winogrande), covering almost half of the gap between the fully abstracted setting and the non-abstracted original model (79.1). This indicates that our contrastive explanations encode a significant amount of information required for commonsense tasks.
Even if the full model does not always use the explanations, these evaluations show that our contrastive explanations contain rich task-relevant knowledge, and suggest that future work might focus on how to better make use of this signal.

\begin{table}[]
    \centering
    \begin{tabular}{lr}
\toprule
Model &	Acc. \\
\midrule
Random & 20.0 \\
Baseline & 	37.2 \\
Self talk & 26.9 \\
Contrastive (V) & 38.1 \\
Contrastive (MM) & 37.4 \\
\midrule
\citet{banerjee2020self} &	38.8 \\
\bottomrule
    \end{tabular}
    \caption{Zero-shot test performance on CommonsenseQA for baselines as well as contrastive models which ensemble fact/foil pairs by voting (V) and maximum margin (MM). The best reported unsupervised performance \cite{banerjee-baral-2020-self} uses ConceptNet, which was used to construct the dataset.} \label{tab:commonsense_qa} 
\end{table}

\subsection{Generalizability of Prompts}
The set of contrastive prompts used in our framework are curated from an in-house analysis of training instances from Winogrande and PIQA datasets. To determine the generalizability of these prompts for other commonsense reasoning tasks, we also experiment with the CommonsenseQA dataset \cite{talmor-etal-2019-commonsenseqa}, which consists of multiple-choice questions created over  ConceptNet -- ``Where on a river can you hold a cup upright to catch water on a sunny day? a) waterfall, b) bridge, c) valley, d)  pebble, e) mountain". Since there are more than two answer choices to contrast, we convert each instance into 10 pairwise (binary) classification instances. Contrastive explanations are generated for each pairwise decision in the zero-shot setting, similar to Winograd and PIQA datasets. To choose the final answer, we consider two inference procedures: (a) \textit{Vote:} The answer that receives the maximum number of votes across all binary classification instances is selected, and (b) \textit{Maximum Margin:} The choice with the maximum difference (margin) between answer likelihoods for any binary classification instance is selected. 
In Table \ref{tab:commonsense_qa}, we observe that self-talk  significantly hurts performance for this dataset. On the other hand, contrastive explanations are found to be useful and approach the zero-shot performance of the state-of-the-art, which uses ConceptNet \cite{banerjee-baral-2020-self}.
These results demonstrate that the set of contrastive prompts are generalizable to other commonsense reasoning datasets, and that while our contrastive prompts are limited to contrasting two answer choices at a time, the framework can be easily extended to tasks with multiple foils.

\section{Conclusion}
We show it is possible to prompt pretrained language models (PLMs) to generate contrastive explanations of their reasoning patterns, inspired by explanations that humans naturally provide for their reasoning. 
Conditioning model decisions on these explanations improves performance on two commonsense reasoning benchmarks, and humans judge the explanations to be highly relevant and helpful in comparison to prior work.
We also showed how contrastive explanations can facilitate in-depth evaluations of faithfulness by flipping or abstracting the fact and foil, finding that our explanations encode a significant amount of information relevant to the classification decision, and in many cases models rely on the contrast in the expected way.
While we have shown that our method is flexible enough to apply to multiple-choice commonsense tasks with many foils, leveraging contrastive reasoning in a wider variety of open-ended tasks remains an exciting challenge for future work.


\section*{Acknowledgements}
This research was supported by ONR N00014-18-1-
2826, DARPA N66001-19-2-403, ARO W911NF16-1-0121 and NSF IIS-1252835, IIS-1562364, an
Allen Distinguished Investigator Award, and the
Sloan Fellowship. We thank Vered Shwartz, Mandar Joshi, Divyansh Kaushik, H2Lab members, UW NLP and the
anonymous reviewers for their helpful comments
and suggestions.

\bibliography{anthology,acl2021}
\bibliographystyle{acl_natbib}

\appendix

\section{Generating Contrastive Templates \label{apdx:contrastive_templates_all}}

\begin{table*}[h!]
    \centering
    \small
    \begin{tabular}{l}
    \toprule
    Winogrande \\
    \midrule
    Ian volunteered to eat Dennis's menudo after already having a bowl because \_\_ despised eating \\
    $a_1$ : Ian \\
    $a_2$ : Dennis \\
    $a_0$ : he \\
    $c_{a_0}$ : Ian volunteered to eat Dennis's menudo after already having a bowl because \underline{he} despised eating \\
    $c_{a_1}$ : Ian volunteered to eat Dennis's menudo after already having a bowl because \underline{Ian} despised eating \\
    $c_{a_2}$ : Ian volunteered to eat Dennis's menudo after already having a bowl because \underline{Dennis} despised eating \\
    \midrule
    PIQA (difference between answers is 1-2 words) \\
    \midrule
    To prepare carrots before cooking with them, you can \\
    $a_1$ : Run them in the sink under boiling water \\
    $a_2$ : Run them in the sink under cold water \\
    $a_0$ : boiling water \underline{or} cold water \\
    $c_{a_0}$ : To prepare carrots before cooking with them, you can run them in the sink under \underline{boiling water }\\\underline{or cold water}\\
    $c_{a_1}$ : To prepare carrots before cooking with them, you can run them in the sink under \underline{boiling water}\\
    $c_{a_2}$ : To prepare carrots before cooking with them, you can run them in the sink under \underline{cold water}\\
    \midrule
    PIQA (difference between answers is larger) \\
    \midrule
    To prevent gunk buildup in cup holders of a car, \\
    $a_1$ : place coffee filters inside of the cup holders.\\
    $a_2$ : pour a thin layer of oil into the cup holders. \\
    $a_0$ : place coffee filters inside of the cup holders \underline{or} pour a thin layer of oil into the cup holders. \\
    $c_{a_0}$ : To prevent gunk buildup in cup holders of a car, \underline{place coffee filters inside of the cup holders or} \\ \underline{pour a thin layer of oil into the cup holders }\\
    $c_{a_1}$ : To prevent gunk buildup in cup holders of a car, \underline{place coffee filters inside of the cup holders}\\
    $c_{a_2}$ : To prevent gunk buildup in cup holders of a car, \underline{pour a thin layer of oil into the cup holders }\\
    \bottomrule
    \end{tabular}
    \caption{Examples of Winogrande and PIQA, with fact, foil , neutral answer and respective substituted contexts used in our approach for prompting the explainer PLM or computing answer likelihood.}
    \label{tab:transformation_example}
\end{table*}

\begin{table*}[h!]
    \centering
    \small
    \begin{tabular}{l}
    \toprule
    Original Input: The geese prefer to nest in the fields rather than the forests because in the \_ predators \\are more hidden. \\
    \midrule
    \bf (i) Context-Only \\
    Input to task model: The geese prefer to nest in the \texttt{<mask1>} rather than the \texttt{<mask2>} because in the \_ predators \\ are more hidden. \\
    \midrule
    \bf (ii) Fully Abstracted \\
    Input to explainer: The geese prefer to nest in the \texttt{<mask1>} rather than the \texttt{<mask2>} because in the \_ predators \\ are more hidden. \\
    Generated Explanation: \texttt{<mask1>} is smaller than \texttt{<mask2>} \\
    Input to task model: The geese prefer to nest in the \texttt{<mask1>} rather than the \texttt{<mask2>} because in the \_ predators \\ are more hidden. \texttt{<mask1>} is smaller than \texttt{<mask2>} \\
    \midrule
    \bf (iii) Abstraction after Explanation \\
    Input to explainer: The geese prefer to nest in the fields rather than the forests because in the \_  \\ predators are more hidden. \\
    Generated Explanation: Forests have more predators than fields \\
    Input to task model: The geese prefer to nest in the \texttt{<mask1>} rather than the \texttt{<mask2>} because in the \_ predators \\ are more hidden. \texttt{<mask2>} have more predators than \texttt{<mask1>} \\
    \bottomrule
    \end{tabular}
    \caption{Input to Explainer and Task model for Abstractive Evaluation}
    \label{tab:abstractive_example}
\end{table*}

Table \ref{tab:templates_all} shows the complete list of contrastive patterns used in our work, categorized under different types of attributes/properties. For templates with no place holders for the explainer to fill out, we only replace placeholders for answers (fact and foil). 
Table \ref{tab:transformation_example} lists $a_0, a_1, a_2$, $c_{a_0}$, $c_{a_1}, c_{a_2}$ for different examples in Winogrande and PIQA to explain dataset specific transformations made by our approach. \\
{\it Detection of $P$, $Q$}: For WSC, the fact and foil are typically 1-word nouns. However, they may by qualified in the context and these qualifiers are important for contrasting. For instance, in the WSC example ``She remembered how annoying it is to dust her wood chair so she bought a plastic table instead.", chair and table are the fact and foil. Their qualifiers wood and plastic are important for the construction of the contrast. Hence we retain these qualifiers when preparing prompts for the explainer PLM. Similarly, for PIQA, qualifiers are retained in the prompts.

{\it Case filtering}: We detect case of entities and accordingly filter out templates that are ungrammatical depending on whether the fact and foil are singular/plural. 

{\it Template filtering for WSC}: For examples that do not contain PERSON named entities, we do not include prompts about personal characteristics. Similarly, for examples that contain PERSON named entities, Temporal, Usecase and some spatial patterns were left out.

{\it Template filtering for PIQA}:  We remove all templates about personal characteristics as this dataset deals with physical commonsense.\\


\section{Human Evaluation \label{apdx:human_evaluation}}
The annotation task was carried out in Amazon
Mechanical Turk, following \cite{shwartz2020unsupervised}. To ensure the quality of annotations, workers were required to be located in the
US, UK, or Canada, and have a 99\% approval rate
for at least 5000 prior tasks. Annotators were paid $.30\$$ per HIT to ensure participants get approximately \$15/hr if they are doing the task. Annotation were aggregated from 3 workers using majority vote. The annotations yielded moderate levels of agreement, with
Fleiss Kappa $\kappa$ = 0.43 \cite{landis1977measurement}.

\section{Hyperparameters \label{apdx:hyperparameters}}
\paragraph{Explainer PLM} 
For T5 we use special symbols \texttt{<extra\_id\_0>} and \texttt{<extra\_id\_1>} in place of the blanks (\_) in our templates. We observe that T5 is able to replace these tokens with multi-word phrases.
For BART, we substitute blanks with a sequence with four \texttt{[MASK]} tokens to encourage generating multiple words. BART can choose to delete a \texttt{[MASK]} token during generation.
Top-K decoding was done with a beam size of 200 and a maximum output sequence length of 20 for T5 models and 100 for BART. This is because both T5 is pre-trained to in-fill by only generating missing phrases while BART is pre-trained to decode the entire input with missing phrases filled in. We used early stopping for BART. 

\paragraph{Task PLM} Task PLM was finetuned for 20 epochs, using BertAdam optimizer with a learning rate of $2e
-5$, batch size of $8$, and dropout of $0.1$, following \cite{latcinnik2020explaining}.

\paragraph{Self-Talk}
\cite{shwartz2020unsupervised} generate multiple clarification questions conditioned on the context, by 1) concatenating one of several question prefixes to the input prompt or question; and 2) generating 5 questions for each prefix using Nucleus sampling with
$p=0.2$, i.e., sampling from the top 20\% tokens\cite{holtzman2019curious}  limiting the question
length to up to 6 tokens excluding the prefix. For each well-formed question, they generate multiple answers using a
similar method. They limit the answer length
to 10 generated tokens, and use Nucleus sampling
with $p=0.5$. \citet{shwartz2020unsupervised} only condition task prediction on a single clarification question and answer pair that increases the model's belief of a certain
answer choice. Thus, the score of each answer
choice is selected as the score of the text containing the clarification that most supports it, i.e.,
whose combination with it yields maximum language model likelihood.

\paragraph{Unconstrained Generation}
For unconstrained explanation baseline, maximum output sequence length was set to 20 and beam size for beam decoding was set to 200. Again we use early stopping.

\begin{table*}[h!]
    \centering
    \begin{small}
    \begin{tabular}{l|l}
    \toprule
    \bf Complete list of Contrastive Prompt Templates & Commonsense Task/Instance Type \\
    \midrule
\bf Temporal: & PIQA (Consists of events) \\
OPT1 happened before/after OPT2 \\
OPT1 takes longer than OPT2 \\
OPT1 takes longer to \_ than OPT2 \\
OPT1 happened for a longer time than OPT2 \\
\midrule
\bf Personal Characteristics: & WSC \\
OPT1 likes \_ while OPT2 likes \_ & (if PERSON entity tag is detected)\\
OPT1 likes \_ while OPT2 does not like \_ \\
OPT1 likes to \_ while OPT2 likes to \_ \\
OPT1 likes to \_ while OPT2 does not like to \_ \\
OPT1 prefers \_ while OPT2 prefers \_ \\
OPT1 prefers \_ while OPT2 does not prefer \_ \\
OPT1 prefers to \_ while OPT2 prefers to \_ \\
OPT1 prefers to \_ while OPT2  does not prefer to \_ \\
OPT1 thinks \_ while OPT2 thinks \_ \\
OPT1 thinks \_ while OPT2 does not thinks \_ \\
\midrule
\bf Object Characteristic: & WSC and PIQA \\
OPT1 is/are smaller than OPT2 & \\
OPT1 is/are larger than OPT2 \\
OPT1 is/are slower than OPT2 \\
OPT1 is/are faster than OPT2 \\
OPT1 is \_ than OPT2 \\
OPT1 are \_ than OPT2 \\
OPT1 is \_ while OPT2 is \_ \\
OPT1 is \_ but OPT2 is \_ \\
OPT1 is \_ however OPT2 is \_ \\
OPT1 are \_ while OPT2 are \_ \\
OPT1 are \_ but OPT2 are \_ \\
OPT1 are \_ however OPT2 are \_ \\
OPT1 has \_ while/but/however OPT2 has/does not have \_ \\
OPT1 have \_ while/but/however OPT2 have/do not have \_ \\
OPT1 is made of/to \_ however OPT2 is made of/to \_ \\
OPT1 is made of/to \_ while OPT2 is made of/to \_ \\
\midrule
\bf Spatial: & WSC and PIQA\\
OPT1 is above OPT2 \\
OPT1 is below OPT2 \\
OPT1 is to the right of OPT2 \\
OPT1 is to the left of OPT2 \\
OPT1 is inside OPT2 \\
OPT1 is outside OPT2 \\
\_ is closer to OPT1 and father away from OPT2 \\
OPT1 is closer to \_ while OPT2 is father away from \_ \\
\midrule
\bf Usecase: & WSC(No PERSON entity) and PIQA \\
OPT1 can \_  while OPT2 can/cannot \_ \\
OPT1 is/can be used for OPT2 \\
OPT1 is/can be  used to do OPT2 \\
OPT1 is/can be  used for \_ but OPT2 cannot \\
OPT1 is/can be  used for \_ while OPT2 is used for \_ \\
OPT1 is/can be s used for \_ but OPT2 is used for \_ \\
OPT1 is/can be used to \_ while OPT2 is used to \_ \\
OPT1 is/can be used to \_ but OPT2 is used to \_ \\
\midrule 
\bf Causes: & WSC (No PERSON entity) and PIQA\\
OPT1 has \_ because \_ while OPT2 is \_ because \_ \\
OPT1 can cause \_ while OPT2 causes/results in \_ \\
Since \_ it can OPT1 but not OPT2 \\
Since \_ it can OPT1 but because it is not \_ it can't OPT2 \\
\midrule
\bf Miscellaneous: & WSC (No PERSON entity) and PIQA\\
\_ can be OPT1 but cannot be OPT2 \\
OPT1 means to \_ while OPT2 means to \_ \\
OPT1 is defined as \_ while OPT2 is defined as \_ \\
\_ OPT1 \_ OPT2 \\
\_ OPT1 but not OPT2 \\
OPT1 exists while an OPT2 doesn't \\ 
    \bottomrule
    \end{tabular}
    \end{small}
    \caption{Complete list of contrastive patterns used in this work.}
    \label{tab:templates_all}
\end{table*}


\end{document}